\documentclass[11pt]{amsart}

\usepackage{amssymb}
\usepackage{amsmath}
\usepackage{amsthm}
\usepackage{graphicx}
\usepackage{bm}
\usepackage[usenames,dvipsnames]{xcolor}

\usepackage{subcaption}
\usepackage{float}
\usepackage{color}
\usepackage{xurl}
\usepackage{hyperref}
\hypersetup{
  colorlinks,
  citecolor= Brown,
  linkcolor= link,
  urlcolor= link
}
\usepackage{enumitem}

\usepackage{tikz}
\usepackage{pgfplots}
\usepackage[export]{adjustbox}
\usetikzlibrary{spy}

\pgfplotsset{compat=1.18}

\pgfplotsset{
    colormap={bw}{
        rgb255(0cm)=(0,0,0);  
        rgb255(1cm)=(255,255,255)  
    }
}

\pgfplotsset{
 colormap={magma_custom}{
rgb=(0.001462,0.000466,0.013866)
rgb=(0.001462,0.000466,0.013866)
rgb=(0.002258,0.001295,0.018331)
rgb=(0.003279,0.002305,0.023708)
rgb=(0.004512,0.00349,0.029965)
rgb=(0.00595,0.004843,0.03713)
rgb=(0.007588,0.006356,0.044973)
rgb=(0.009426,0.008022,0.052844)
rgb=(0.011465,0.009828,0.06075)
rgb=(0.013708,0.011771,0.068667)
rgb=(0.016156,0.01384,0.076603)
rgb=(0.018815,0.016026,0.084584)
rgb=(0.018815,0.016026,0.084584)
rgb=(0.021692,0.01832,0.09261)
rgb=(0.024792,0.020715,0.100676)
rgb=(0.028123,0.023201,0.108787)
rgb=(0.031696,0.025765,0.116965)
rgb=(0.03552,0.028397,0.125209)
rgb=(0.039608,0.03109,0.133515)
rgb=(0.04383,0.03383,0.141886)
rgb=(0.048062,0.036607,0.150327)
rgb=(0.05232,0.039407,0.158841)
rgb=(0.056615,0.04216,0.167446)
rgb=(0.060949,0.044794,0.176129)
rgb=(0.060949,0.044794,0.176129)
rgb=(0.06533,0.047318,0.184892)
rgb=(0.069764,0.049726,0.193735)
rgb=(0.074257,0.052017,0.20266)
rgb=(0.078815,0.054184,0.211667)
rgb=(0.083446,0.056225,0.220755)
rgb=(0.088155,0.058133,0.229922)
rgb=(0.092949,0.059904,0.239164)
rgb=(0.097833,0.061531,0.248477)
rgb=(0.102815,0.06301,0.257854)
rgb=(0.107899,0.064335,0.267289)
rgb=(0.107899,0.064335,0.267289)
rgb=(0.113094,0.065492,0.276784)
rgb=(0.118405,0.066479,0.286321)
rgb=(0.123833,0.067295,0.295879)
rgb=(0.12938,0.067935,0.305443)
rgb=(0.135053,0.068391,0.315)
rgb=(0.140858,0.068654,0.324538)
rgb=(0.146785,0.068738,0.334011)
rgb=(0.152839,0.068637,0.343404)
rgb=(0.159018,0.068354,0.352688)
rgb=(0.165308,0.067911,0.361816)
rgb=(0.171713,0.067305,0.370771)
rgb=(0.171713,0.067305,0.370771)
rgb=(0.178212,0.066576,0.379497)
rgb=(0.184801,0.065732,0.387973)
rgb=(0.19146,0.064818,0.396152)
rgb=(0.198177,0.063862,0.404009)
rgb=(0.204935,0.062907,0.411514)
rgb=(0.211718,0.061992,0.418647)
rgb=(0.218512,0.061158,0.425392)
rgb=(0.225302,0.060445,0.431742)
rgb=(0.232077,0.059889,0.437695)
rgb=(0.238826,0.059517,0.443256)
rgb=(0.238826,0.059517,0.443256)
rgb=(0.245543,0.059352,0.448436)
rgb=(0.25222,0.059415,0.453248)
rgb=(0.258857,0.059706,0.45771)
rgb=(0.265447,0.060237,0.46184)
rgb=(0.271994,0.060994,0.46566)
rgb=(0.278493,0.061978,0.46919)
rgb=(0.284951,0.063168,0.472451)
rgb=(0.291366,0.064553,0.475462)
rgb=(0.29774,0.066117,0.478243)
rgb=(0.304081,0.067835,0.480812)
rgb=(0.310382,0.069702,0.483186)
rgb=(0.310382,0.069702,0.483186)
rgb=(0.316654,0.07169,0.48538)
rgb=(0.322899,0.073782,0.487408)
rgb=(0.329114,0.075972,0.489287)
rgb=(0.335308,0.078236,0.491024)
rgb=(0.341482,0.080564,0.492631)
rgb=(0.347636,0.082946,0.494121)
rgb=(0.353773,0.085373,0.495501)
rgb=(0.359898,0.087831,0.496778)
rgb=(0.366012,0.090314,0.49796)
rgb=(0.372116,0.092816,0.499053)
rgb=(0.372116,0.092816,0.499053)
rgb=(0.378211,0.095332,0.500067)
rgb=(0.384299,0.097855,0.501002)
rgb=(0.390384,0.100379,0.501864)
rgb=(0.396467,0.102902,0.502658)
rgb=(0.402548,0.10542,0.503386)
rgb=(0.408629,0.10793,0.504052)
rgb=(0.414709,0.110431,0.504662)
rgb=(0.420791,0.11292,0.505215)
rgb=(0.426877,0.115395,0.505714)
rgb=(0.432967,0.117855,0.50616)
rgb=(0.439062,0.120298,0.506555)
rgb=(0.439062,0.120298,0.506555)
rgb=(0.445163,0.122724,0.506901)
rgb=(0.451271,0.125132,0.507198)
rgb=(0.457386,0.127522,0.507448)
rgb=(0.463508,0.129893,0.507652)
rgb=(0.46964,0.132245,0.507809)
rgb=(0.47578,0.134577,0.507921)
rgb=(0.481929,0.136891,0.507989)
rgb=(0.488088,0.139186,0.508011)
rgb=(0.494258,0.141462,0.507988)
rgb=(0.500438,0.143719,0.50792)
rgb=(0.506629,0.145958,0.507806)
rgb=(0.506629,0.145958,0.507806)
rgb=(0.512831,0.148179,0.507648)
rgb=(0.519045,0.150383,0.507443)
rgb=(0.52527,0.152569,0.507192)
rgb=(0.531507,0.154739,0.506895)
rgb=(0.537755,0.156894,0.506551)
rgb=(0.544015,0.159033,0.506159)
rgb=(0.550287,0.161158,0.505719)
rgb=(0.556571,0.163269,0.50523)
rgb=(0.562866,0.165368,0.504692)
rgb=(0.569172,0.167454,0.504105)
rgb=(0.569172,0.167454,0.504105)
rgb=(0.57549,0.16953,0.503466)
rgb=(0.581819,0.171596,0.502777)
rgb=(0.588158,0.173652,0.502035)
rgb=(0.594508,0.175701,0.501241)
rgb=(0.600868,0.177743,0.500394)
rgb=(0.607238,0.179779,0.499492)
rgb=(0.613617,0.181811,0.498536)
rgb=(0.620005,0.18384,0.497524)
rgb=(0.626401,0.185867,0.496456)
rgb=(0.632805,0.187893,0.495332)
rgb=(0.639216,0.189921,0.49415)
rgb=(0.639216,0.189921,0.49415)
rgb=(0.645633,0.191952,0.49291)
rgb=(0.652056,0.193986,0.491611)
rgb=(0.658483,0.196027,0.490253)
rgb=(0.664915,0.198075,0.488836)
rgb=(0.671349,0.200133,0.487358)
rgb=(0.677786,0.202203,0.485819)
rgb=(0.684224,0.204286,0.484219)
rgb=(0.690661,0.206384,0.482558)
rgb=(0.697098,0.208501,0.480835)
rgb=(0.703532,0.210638,0.479049)
rgb=(0.703532,0.210638,0.479049)
rgb=(0.709962,0.212797,0.477201)
rgb=(0.716387,0.214982,0.47529)
rgb=(0.722805,0.217194,0.473316)
rgb=(0.729216,0.219437,0.471279)
rgb=(0.735616,0.221713,0.46918)
rgb=(0.742004,0.224025,0.467018)
rgb=(0.748378,0.226377,0.464794)
rgb=(0.754737,0.228772,0.462509)
rgb=(0.761077,0.231214,0.460162)
rgb=(0.767398,0.233705,0.457755)
rgb=(0.773695,0.236249,0.455289)
rgb=(0.773695,0.236249,0.455289)
rgb=(0.779968,0.238851,0.452765)
rgb=(0.786212,0.241514,0.450184)
rgb=(0.792427,0.244242,0.447543)
rgb=(0.798608,0.24704,0.444848)
rgb=(0.804752,0.249911,0.442102)
rgb=(0.810855,0.252861,0.439305)
rgb=(0.816914,0.255895,0.436461)
rgb=(0.822926,0.259016,0.433573)
rgb=(0.828886,0.262229,0.430644)
rgb=(0.834791,0.26554,0.427671)
rgb=(0.834791,0.26554,0.427671)
rgb=(0.840636,0.268953,0.424666)
rgb=(0.846416,0.272473,0.421631)
rgb=(0.852126,0.276106,0.418573)
rgb=(0.857763,0.279857,0.415496)
rgb=(0.86332,0.283729,0.412403)
rgb=(0.868793,0.287728,0.409303)
rgb=(0.874176,0.291859,0.406205)
rgb=(0.879464,0.296125,0.403118)
rgb=(0.884651,0.30053,0.400047)
rgb=(0.889731,0.305079,0.397002)
rgb=(0.8947,0.309773,0.393995)
rgb=(0.8947,0.309773,0.393995)
rgb=(0.899552,0.314616,0.391037)
rgb=(0.904281,0.31961,0.388137)
rgb=(0.908884,0.324755,0.385308)
rgb=(0.913354,0.330052,0.382563)
rgb=(0.917689,0.3355,0.379915)
rgb=(0.921884,0.341098,0.377376)
rgb=(0.925937,0.346844,0.374959)
rgb=(0.929845,0.352734,0.372677)
rgb=(0.933606,0.358764,0.370541)
rgb=(0.937221,0.364929,0.368567)
rgb=(0.940687,0.371224,0.366762)
rgb=(0.940687,0.371224,0.366762)
rgb=(0.944006,0.377643,0.365136)
rgb=(0.94718,0.384178,0.363701)
rgb=(0.95021,0.39082,0.362468)
rgb=(0.953099,0.397563,0.361438)
rgb=(0.955849,0.4044,0.360619)
rgb=(0.958464,0.411324,0.360014)
rgb=(0.960949,0.418323,0.35963)
rgb=(0.96331,0.42539,0.359469)
rgb=(0.965549,0.432519,0.359529)
rgb=(0.967671,0.439703,0.35981)
rgb=(0.967671,0.439703,0.35981)
rgb=(0.96968,0.446936,0.360311)
rgb=(0.971582,0.45421,0.36103)
rgb=(0.973381,0.46152,0.361965)
rgb=(0.975082,0.468861,0.363111)
rgb=(0.97669,0.476226,0.364466)
rgb=(0.97821,0.483612,0.366025)
rgb=(0.979645,0.491014,0.367783)
rgb=(0.981,0.498428,0.369734)
rgb=(0.982279,0.505851,0.371874)
rgb=(0.983485,0.51328,0.374198)
rgb=(0.984622,0.520713,0.376698)
rgb=(0.984622,0.520713,0.376698)
rgb=(0.985693,0.528148,0.379371)
rgb=(0.9867,0.535582,0.38221)
rgb=(0.987646,0.543015,0.38521)
rgb=(0.988533,0.550446,0.388365)
rgb=(0.989363,0.557873,0.391671)
rgb=(0.990138,0.565296,0.395122)
rgb=(0.990871,0.572706,0.398714)
rgb=(0.991558,0.580107,0.402441)
rgb=(0.992196,0.587502,0.406299)
rgb=(0.992785,0.594891,0.410283)
rgb=(0.992785,0.594891,0.410283)
rgb=(0.993326,0.602275,0.41439)
rgb=(0.993834,0.609644,0.418613)
rgb=(0.994309,0.616999,0.42295)
rgb=(0.994738,0.62435,0.427397)
rgb=(0.995122,0.631696,0.431951)
rgb=(0.99548,0.639027,0.436607)
rgb=(0.99581,0.646344,0.441361)
rgb=(0.996096,0.653659,0.446213)
rgb=(0.996341,0.660969,0.45116)
rgb=(0.99658,0.668256,0.456192)
rgb=(0.996775,0.675541,0.461314)
rgb=(0.996775,0.675541,0.461314)
rgb=(0.996925,0.682828,0.466526)
rgb=(0.997077,0.690088,0.471811)
rgb=(0.997186,0.697349,0.477182)
rgb=(0.997254,0.704611,0.482635)
rgb=(0.997325,0.711848,0.488154)
rgb=(0.997351,0.719089,0.493755)
rgb=(0.997351,0.726324,0.499428)
rgb=(0.997341,0.733545,0.505167)
rgb=(0.997285,0.740772,0.510983)
rgb=(0.997228,0.747981,0.516859)
rgb=(0.997228,0.747981,0.516859)
rgb=(0.997138,0.75519,0.522806)
rgb=(0.997019,0.762398,0.528821)
rgb=(0.996898,0.769591,0.534892)
rgb=(0.996727,0.776795,0.541039)
rgb=(0.996571,0.783977,0.547233)
rgb=(0.996369,0.791167,0.553499)
rgb=(0.996162,0.798348,0.55982)
rgb=(0.995932,0.805527,0.566202)
rgb=(0.99568,0.812706,0.572645)
rgb=(0.995424,0.819875,0.57914)
rgb=(0.995131,0.827052,0.585701)
rgb=(0.995131,0.827052,0.585701)
}
}

\newcommand{\customcolorbar}[6]{
    \pgfplotscolorbardrawstandalone[
        colormap name=#1, 
        colorbar horizontal,
        point meta min=#2, 
        point meta max=#3, 
        colorbar style={
            width=#4, 
            height=0.2cm,
            xtick={(#2 + #3)/2}, 
            xticklabel style={
                /pgf/number format/fixed, 
                font=\scriptsize, 
                scaled ticks=false 
            },
            xtick pos=bottom, 
            extra x ticks={#2, #3}, 
            extra x tick labels={#5#2, #6#3} 
        }
    ]
}

\definecolor{link}{rgb}{0.18,0.25,0.63}
\definecolor{myred}{rgb}{0.7,0.25,0.2}
\definecolor{mygray}{rgb}{0.8,0.8,0.8}
\usepackage[top=3.3cm,bottom=3.3cm,left=3.5cm,right=3.5cm]{geometry}

\numberwithin{equation}{section}

\newcommand{\bv}{\mathrm{BV}}
\newcommand{\tv}{\mathrm{TV}}
\newcommand{\tgv}{\mathrm{TGV}}
\newcommand{\RR}{\mathbb{R}}
\newcommand{\om}{\Omega}
\newcommand{\LL}{\mathrm{\Lambda}}

\definecolor{myred}{rgb}{0.8,0.25,0.2}
\definecolor{mygreen}{RGB}{0, 153, 0}
\definecolor{myblue}{RGB}{0, 153, 255}
\definecolor{myorange}{RGB}{255, 153, 0}

\newcommand{\cmmnt}[1]{}

\makeatletter
\g@addto@macro{\endabstract}{\@setabstract}
\newcommand{\authorfootnotes}{\renewcommand\thefootnote{\@fnsymbol\c@footnote}}%
\makeatother

\newtheorem{proposition}{Proposition}

\begin{document}
\definecolor{link}{rgb}{0,0,0}
\definecolor{mygrey}{rgb}{0.34,0.34,0.34}
\def\blue #1{{\color{blue}#1}}

 \begin{center}
 \large

   \textbf{Deep unrolling for learning optimal spatially varying regularisation parameters  for Total Generalised Variation} \par \bigskip \bigskip
   \normalsize
  \textsc{Thanh Trung Vu}\textsuperscript{$\,1,$$2$}, \textsc{Andreas Kofler}\textsuperscript{$\,3$},  \textsc{Kostas Papafitsoros}\textsuperscript{$\,2$}
\let\thefootnote\relax\footnote{
\textsuperscript{$1$}University of Cambridge, UK
}
\let\thefootnote\relax\footnote{
\textsuperscript{$2$}School of Mathematical Sciences, Queen Mary University of London, UK}

\let\thefootnote\relax\footnote{
\textsuperscript{$3$}Physikalisch-Technische Bundesanstalt (PTB), Braunschweig and Berlin, Germany}

\let\thefootnote\relax\footnote{
\hspace{3.2pt}Emails: \href{mailto:ttv22@cam.ac.uk}{\nolinkurl{ttv22@cam.ac.uk}},
 \href{mailto: andreas.kofler@ptb.de}{\nolinkurl{andreas.kofler@ptb.de}},
 \href{mailto: k.papafitsoros@qmul.ac.uk}{\nolinkurl{k.papafitsoros@qmul.ac.uk}}
}
\end{center}
\vspace{-0.8cm}

\begin{abstract}
We extend a recently introduced deep unrolling framework for learning spatially varying regularisation parameters in inverse imaging problems to the case of Total Generalised Variation (TGV). The framework combines a deep convolutional neural network (CNN) inferring the two spatially varying TGV  parameters with an unrolled algorithmic scheme that solves the corresponding variational problem. The two subnetworks are jointly trained end-to-end in a supervised fashion and as such the CNN learns to compute those parameters that drive the reconstructed images as close as possible to the ground truth. Numerical results in image denoising and MRI reconstruction show a significant qualitative and quantitative improvement compared to the best TGV scalar parameter case as well as to other approaches employing spatially varying parameters computed by unsupervised methods.
We also observe that the inferred spatially varying parameter maps have a consistent structure near the image edges, asking for further theoretical investigations. In particular, the parameter that weighs the first-order TGV term has a triple-edge structure with alternating high-low-high values whereas the one that weighs the second-order term attains small values in a large neighbourhood around the edges.\\

\noindent
\textbf{Keywords:} Spatially Varying Regularisation Parameters  $\cdot$ Inverse Problems $\cdot$  Total Generalised Variation $\cdot$  Denoising $\cdot$  Magnetic Resonance Imaging $\cdot$  Neural networks $\cdot$  Unrolling

\end{abstract}


\definecolor{link}{rgb}{0.18,0.25,0.63}

\section{Introduction}

In inverse imaging problems, variational regularisation  problems of the type
\begin{equation}\label{intro:general_min}
\min_{u\in X} \; \mathcal{D}(Au,f)+ \mathcal{R}(u;\mathrm{\LL})
\end{equation}
are widely used to compute an estimation $u\in X$ of some ground truth imaging data $u_{\mathrm{true}}$, $X$ being a Banach space,  given data $f$ that satisfy the equation
\begin{equation}\label{intro:general_eq}
f=Au_{\mathrm{true}}+\eta.
\end{equation}
Here, $A$ denotes the forward operator of the, typically ill-posed, inverse problem and $\eta$ is a random noise component.
Solving \eqref{intro:general_min} using spatially varying regularisation parameters $\LL$, instead of scalar ones,
has been the subject of many  works, see \cite{Calatroni_bilevellearning,bilevel_handbook,Pragliola_SIAMreview} and the references therein.
The goal  is to compute and subsequently use a regularisation parameter  $\mathrm{\Lambda}: \Omega \to (\mathbb{R}_{+})^{\ell}$ that balances the data fidelity  $\mathcal{D}$ and the -- in general $\ell$ components -- of the regularisation term $\mathcal{R}$ with a different strength at every point $x\in \Omega \subset \mathbb{R}^{d}$ (every pixel in the pixel-domain $\Omega$, in the discrete setting).
In the case where $\mathcal{R}(u)=\mathrm{TV}(u)$, the Total Variation of the (grayscale) function $u: \Omega \to \mathbb{R}$ and assuming Gaussian noise, problem \eqref{intro:general_min} amounts to
\begin{equation}\label{intro:weighted_TV_min}
\min_{u\in X}\; \frac{1}{2}\|Au-f\|_{L^2(\Omega)}^{2} +\int_{\Omega} \LL(x) d|Du|.
\end{equation}
Small values of $\LL: \Omega \to \mathbb{R}_{+}$ impose little local regularity and are thus suitable for preserving detailed parts of the image and edges. On the other hand, high values of $\LL$  impose large regularity and are preferable for smooth, homogeneous areas.

For higher-order extensions of TV, especially those defined in an infimal convolution manner, the role and the interplay of spatially varying regularisation parameters on the resulting image quality and structure are not as straightforward. A prominent example is the Total Generalised Variation (TGV) \cite{TGV}
\begin{equation}\label{intro:weighted_TGV}
\tgv_{\LL_{0}, \LL_{1}}(u):=\min_{w\in \mathrm{BD}(\om)} \int_{\Omega} \LL_{1}(x) d|Du-w| + \int_{\Omega} \LL_{0}(x)d|\mathcal{E}w|,
\end{equation}
where $\LL: \Omega \to (\mathbb{R}_{+})^{2}$, with $\LL=(\LL_{0}, \LL_{1})$. Here, $\mathcal{E}$ denotes the measure that represents the distributional symmetrised gradient of $w\in \mathrm{BD}(\Omega)$, the space of functions of bounded deformations.
The combined values of $\LL_{0}$ and $\LL_{1}$ not only regulate the regularisation intensity but also control the staircasing effect, which is a characteristic limitation of  TV, with suitable values of these parameters promoting piecewise affine structures.

 In contrast to TV, where multiple works have considered computing spatially varying $\LL$ in \eqref{intro:weighted_TV_min}, see aforementioned reviews, there are limited works that focus on the computation of such spatially varying  $\LL_{0}, \LL_{1}$ for TGV.   In \cite{bilevelTGV}, a bilevel unsupervised scheme was used, employing a statistics-based upper level energy.
 The approach produced satisfactory results, albeit with a high computational cost and a need to impose continuity assumptions to $\LL_{0}, \LL_{1}$ for robustness reasons. The latter regularity for the parameter-maps limits their adaptability to the image structure, thus not fully exploiting the potential of spatial adaptivity.

Recently, there has been a series of works focusing on learning regularisation parameters for TV using  neural networks \cite{Afkham_2021,Nekhili_2022,Cuomo_2024}. There, a network is initially suitably trained in a supervised manner to infer regularisation parameters, which in a second phase are used in the corresponding variational problem that is solved as usual. Thus, one exploits the versatility of neural networks to inform a model-based variational regularisation scheme whose solution is interpretable since the regulariser remains handcrafted.

\subsection*{Contribution}
Here, we adapt the approach introduced in \cite{Kofler_2023} to compute spatially varying regularisation maps for TGV. It involves training a network that consists of two subnetworks in a supervised fashion. The first subnetwork is a deep convolutional neural network (CNN), 
that takes as an input the data $f$
and outputs the maps $\LL_{0}, \LL_{1}$. To tie these maps to the variational problem, they are fed into a second  appended  subnetwork, an unrolled PDHG algorithm \cite{chambolle2011first}, that solves the TGV minimisation problem considering the regularisation maps to be fixed. The entire network is trained end-to-end with pairs  $(f^{i}, u_{\mathrm{true}}^{i})$ and thus, the CNN is implicitly trained to output those maps that drive the approximate solution $u^{i}$ of the variational problem as close as possible to $u_{\mathrm{true}}^{i}$.
Since the CNN is expressive enough, given some new data $f^{\mathrm{test}}$, it can produce meaningful parameter-maps adapted to $f^{\mathrm{test}}$. We show that this approach produces far better results than similar approaches in image denoising and MR image reconstruction, significantly boosting the performance of TGV. We also show that in contrast to the TV, the structure of the resulting TGV parameter maps is non-trivial and asks for further theoretical investigation.

\noindent
\emph{Outline}:
In Section \ref{sec:framework}, we set some notation and recall some preliminaries about TV and TGV regularisation with scalar and spatially varying parameter maps. We proceed to describe the deep unrolled PDHG network for learning these maps. In Sections \ref{sec:denoising} and \ref{sec:mri}, we present our numerical results in denoising and MRI reconstruction, respectively, and discuss the structure of the resulting maps near the image edges. We conclude in Section \ref{sec:conclusion}.

\section{The framework}\label{sec:framework}
\subsection{Preliminaries}\label{sec:preliminaries}
We initially define the several notions in function space to recall a few properties that stem from their analysis. As usual, $\om\subset \RR^{d}$ is a bounded, Lipschitz domain and we consider grayscale images, i.e.\ $\RR$-valued.
For positive functions $\LL, \LL_{0}, \LL_{1}: \om\to \RR$ we define the spatially varying TV and TGV functionals as
\begin{align}
\tv_{\LL}(u)&= \int_{\om}\LL(x) d|Du|, \label{TVLLambda}\\
\tgv_{\LL_{0}, \LL_{1}}(u)&= \min_{w\in \mathrm{BD}(\om)}\ \int_{\om}\LL_{1}(x) d|Du-w| + \int_{\om} \LL_{0}(x) d|\mathcal{E}w|. \label{TGVLLambda01}
\end{align}
We  also denote by $\tv_{\lambda}$ and $\tgv_{\lambda_{0}, \lambda_{1}}$ the corresponding  functionals with scalar parameters. 
It holds that when the weights $\LL, \LL_{0}, \LL_{1}$ are bounded, lower semicontinuous functions, bounded away from zero,  the functionals \eqref{TVLLambda} and \eqref{TGVLLambda01} are well-defined for $u\in \bv(\om)$ 
\cite{davoli2023dyadic}. Moreover, the following proposition suggests that continuity of the weights is not essential for well-posedness.

\begin{proposition}\label{prop:existence}
Let $(\mathcal{H}, \|\cdot\|_{\mathcal{H}})$ be a Hilbert space, $f\in \mathcal{H}$ and  $\LL, \LL_{0}, \LL_{1}\in L^{\infty}(\om)$ be lower semicontinuous,  bounded away from zero. Let also $A\in \mathcal{L}(L^{p}(\om), \mathcal{H})$ with $p\in (1, d^{\ast}]$,  $d^\ast=d/(d-1)$ ($d^{\ast}=\infty$ if $d=1$). Then both problems
\begin{align}
&\min_{u\in \bv(\om)}\; \frac{1}{2}\|Au-f\|_{\mathcal{H}}^{2} +\tv_{\LL}(u), \label{weighted_TV_min}\\
&\min_{u\in \bv(\om)}\; \frac{1}{2}\|Au-f\|_{\mathcal{H}}^{2} +\tgv_{\LL_{0}, \LL_{1}}(u),  \label{weighted_TGV_min}
\end{align}
admit a solution in $\bv(\om)$.
\end{proposition}
 \begin{proof}
 The proof readily follows by combining \cite[Thm 3.2, Prop. 5.9]{davoli2023dyadic}, \cite[Prop. 5.1]{structuralTV} and \cite[Thm. 2.11 \& Prop. 5.17]{bredies2020higher} and we omit it here.
 \end{proof}

\noindent
We also recall the following proposition from \cite{Papafitsoros_Valkonen_2015}, which dictates that when the ratio $\lambda_{0}/\lambda_{1}$ is large enough, $\tgv_{\lambda_{0}, \lambda_{1}}$ essentially behaves like $\tv_{\lambda_{1}}$.
\begin{proposition}[from \cite{Papafitsoros_Valkonen_2015}]\label{prop:tgv_large_ratio}
There exists a constant $C>0$ depending only on the domain $\om$ such that if $\lambda_{0}, \lambda_{1}>0$ satisfy $\lambda_{0}/\lambda_{1}>C$ then
\begin{equation}\label{tgv_large_ratio}
\tgv_{\lambda_{0}, \lambda_{1}}(u)=\lambda_{1} |Du-m_{\mathcal{E}}(\nabla u)|(\Omega), \quad \text{for all }u\in \bv(\om),
\end{equation}
where for  $v\in L^{1}(\om,\RR^{d})$ we define
$m_{\mathcal{E}}:=\operatorname{argmin}\,\{\|v-w\|_{L^{1}(\om,\RR^{d})}:\, w\in Ker \mathcal{E}\}$.
Recall that $Ker \mathcal{E}=\{r(x)=Bx+c:\;c\in\mathbb{R}^{d},\,B\in \mathbb{R}^{d\times d} \text{ skew symmetric}\}$.
\end{proposition}

\subsection{U-Net plus unrolled PDHG for TGV}\label{sec:unet_pdhg}

\begin{figure}[t]
\includegraphics[width=\textwidth]{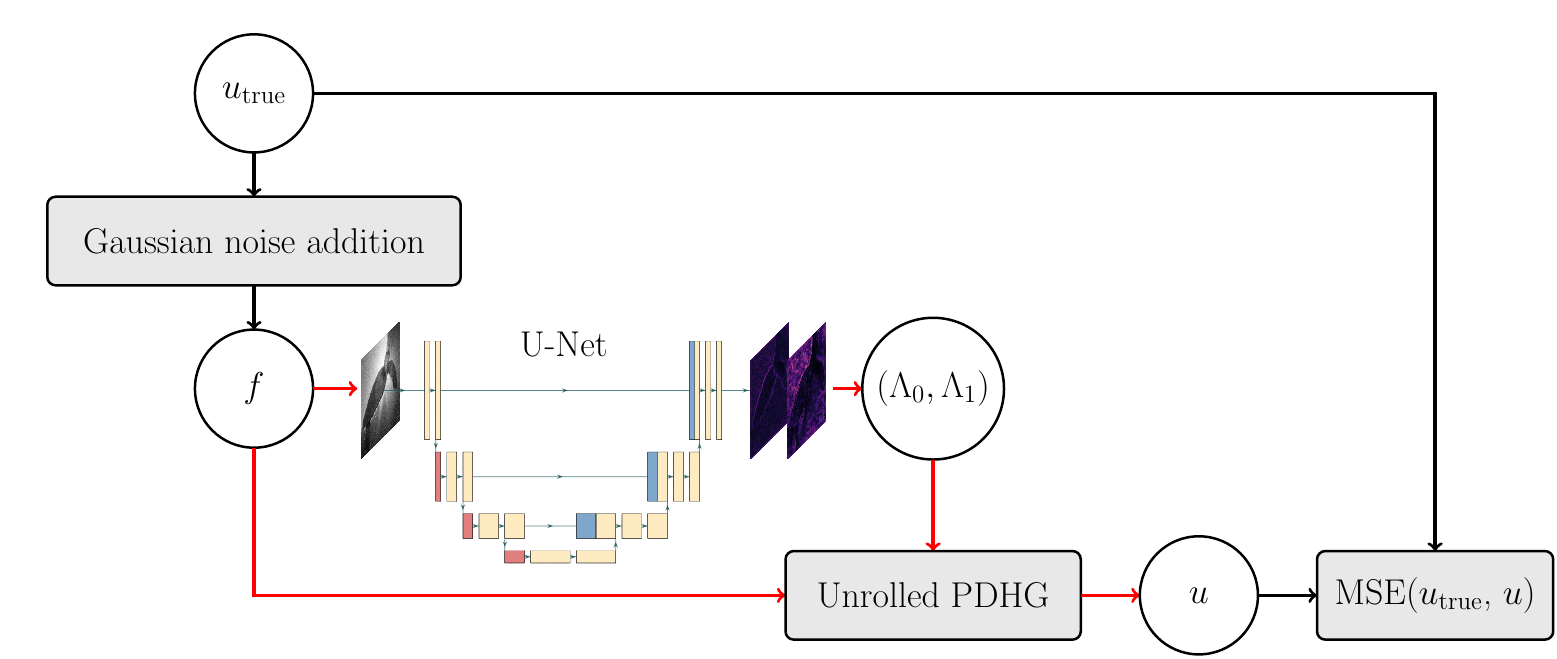}
\caption{Visualisation of the network $\mathcal{N}_{\theta}^{N}: f\mapsto (\mathrm{\Lambda}_{0}, \mathrm{\Lambda}_{1})\mapsto u$ of \eqref{unrolled_network} (red arrows) and its training procedure \eqref{supervised_learning} (black arrows) for the denoising case.}\label{fig:workflow}
\end{figure}

The combined neural network-based algorithm unrolling technique for the estimation of spatially varying $\LL_{0}, \LL_{1}$  that we adopt here follows \cite{Kofler_2023}. Let $u_{n}=S^{n}(\tilde{u}_{0}, f, (\LL_{0}, \LL_{1}), A)$, $n\in\mathbb{N}$ denote the iterates of some algorithm  that solves \eqref{weighted_TGV_min}. That is, $u_{n}\to u^{\ast}$ as $n\to \infty$, where $u^{\ast}$ is a solution of \eqref{weighted_TGV_min}. Here $\tilde{u}_{0}$ denotes a suitable initialisation in the image space e.g.\ $\tilde{u}_{0}=A^{\ast} f$, with $A^{\ast}$ denoting the adjoint of $A$. Next we denote with $\mathrm{NET}_{\theta}:A^{\ast}f \mapsto (\LL_{0}, \LL_{1})$ a deep convolutional neural network with learnable parameters $\theta$ (a U-Net \cite{Ronneberger2015} in our implementations). Then,  for a \emph{fixed} $N\in \mathbb{N}$, we define the overall network
\begin{equation}\label{unrolled_network}
\mathcal{N}_{\theta}^{N}(f)= S^{N}(A^{\ast}(f), f, \mathrm{NET}_{\theta}(A^{\ast}f), A).
\end{equation}
The unrolled network \eqref{unrolled_network} can then be trained end-to-end in a supervised fashion using a dataset of data-ground truth pairs $(f^{i}, u_{\mathrm{true}}^{i})_{i=1}^{M}$, and an appropriate pairwise distance function $l$, 
\begin{equation}\label{supervised_learning}
\min_{\theta}\; \mathrm{Loss}(\theta):= \frac{1}{M} \sum_{i=1}^{M} l(\mathcal{N}_{\theta}^{N}(f^{i}), u_{\mathrm{true}}^{i}),
\end{equation}
see also Figure \ref{fig:workflow} for an illustration of the denoising case.
Hence, the network $\mathrm{NET}_{\theta}$ implicitly learns to output  spatially varying regularisation parameters that force an approximate solution of \eqref{weighted_TGV_min} to be close to the ground truth. Because $\mathrm{NET}_{\theta}$ will be overparameterised and thus expressive enough, it is expected that when the trained $\mathcal{N}_{\theta}^{N}$ is applied to  new unseen data $f^{\mathrm{test}}$ it will still produce suitable data-adaptive parameters even though it does not have access to the  ground truth.
Henceforth, we denote the corresponding networks that follow the above approach by U-TV and U-TGV. We stress that, even though U-TV was introduced in \cite{Kofler_2023}, that work did not include 2D denoising and 2D MRI results, which we provide here for comparison with U-TGV. We finally note that here we used the anisotropic versions of  TV and TGV  with their standard discretisations.

\section{Numerical experiments in image denoising}\label{sec:denoising}

\subsection{Set-up}\label{sec:denoising-set-up}
 \begin{enumerate}[leftmargin=0pt]
 \item[] \emph{The training dataset:} To train the U-TV and U-TGV models for denoising we used the SeaTurtleID2022 dataset \cite{Adam_2024_WACV}, which is originally designed for re-identification tasks. The rationale  was to test among others the model performance when trained on a specific image distribution, here underwater images of sea turtles, and subsequently tested to images outside this distribution. We randomly selected 500 and 50 images for the training and the validation set respectively.
 All images were cropped and rescaled to $512\times512$, converted to grayscale and normalised to $[0,1]$.
 During training, the noisy input images were generated on the fly by corrupting the target images by zero-mean Gaussian noise with standard deviation randomly chosen in $[0, 0.2]$.
 \item[] \emph{The architecture of $\mathrm{NET}_{\theta}$:}
 We used a U-Net architecture \cite{Ronneberger2015} consisting of three encoding and three decoding blocks, and two final convolutional layers, resulting in 1-128-256-512-1024-512-256-128-2 structure (-1 for U-TV). All convolutional layers have a kernel size of $3 \times 3$ and a stride of 1.
 We used the Leaky ReLU as activation function with a negative slope of $0.01$.
 The map $0.1\times \mathrm{softplus(\cdot)}$  was applied to the final layer to guarantee positiveness for the parameters $\LL, \LL_{0}, \LL_{1}$. Overall, the number of trainable parameters $\theta$ was 28,712,577 and 28,712,706 for  U-TV and U-TGV  respectively.
\item[] \emph{The choice of solution algorithm:}
As an algorithm for TV minimisation, we used the standard PDHG algorithm \cite{chambolle2011first}, see also \cite[Alg. 5.1]{Kofler_2023}. Regarding step sizes, we chose $\sigma = \tau = \text{sigmoid}(10) / \sqrt{13}$.  We also used the PDHG algorithm for U-TGV as outlined in \cite{tgvcolour}, setting $\sigma = \tau = 0.29$. In both algorithms, we set the value of extrapolation parameter to $\text{sigmoid}(10)$,  where $\text{sigmoid}(y) = 1 / (1 + \exp(-y))$. We note that these values guarantee the convergence of the algorithms.
\item[] \emph{The choice of $N$:} The number of unrolled PDHG iterations is crucial since it should be large enough to approximate the solution of the variational problem. However, setting it too large potentially unnecessarily increases the computational cost (GPU-memory and time) during training. We set $N=256$ which, according to our observations, achieved a good balance.
\item[] \emph{The training procedure:} We used the Adam optimiser \cite{kingma2014adam}  with
a learning rate of $10^{-4}$, a batch size of 1, and the MSE loss.
The number of training epochs for  U-TV and U-TGV  was 200 and 100 respectively, beyond which, the performance of the models when applied to the validation data no longer improved. Training took approximately 10 and 20 hours for U-TV and U-TGV respectively, on an RTX 4090 (24GB VRAM).\\
The full code for all the experiments can be found in \url{https://github.com/trung-vt/LearningRegularizationParametersForTGV}. 
 \end{enumerate}

\subsection{Results}
\begin{figure}[!t]
    \centering
    \begin{subfigure}[t]{0.24\textwidth}
             \begin{tikzpicture}[spy using outlines={rectangle, white, magnification=2, size=0.9cm, connect spies}]
  	  \node[anchor=south west,inner sep=0]  at (0,0) {
        \includegraphics[width=\textwidth]{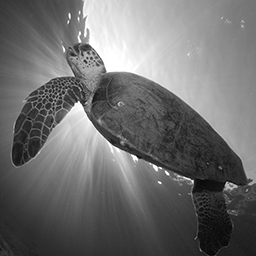}};
   		 \spy on (0.6, 2.3) in node [left] at (1.0, 0.5);
  	  \end{tikzpicture}
        \caption{Ground truth}
    \end{subfigure}
    \begin{subfigure}[t]{0.24\textwidth}
             \begin{tikzpicture}[spy using outlines={rectangle, white, magnification=2, size=0.9cm, connect spies}]
  	  \node[anchor=south west,inner sep=0]  at (0,0) {
        \includegraphics[width=\textwidth]{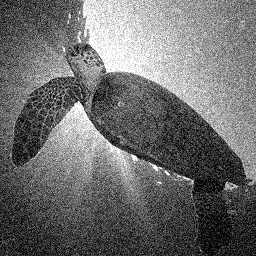}};
   		 \spy on (0.6, 2.3) in node [left] at (1.0, 0.5);
  	  \end{tikzpicture}
        \caption{Noisy, $sd=0.1$  \\ $[19.99,  0.2448]$ \newline}
    \end{subfigure}
    \hfill
    \begin{subfigure}[t]{0.24\textwidth}
             \begin{tikzpicture}[spy using outlines={rectangle, white, magnification=2, size=0.9cm, connect spies}]
  	  \node[anchor=south west,inner sep=0]  at (0,0) {
        \includegraphics[width=\textwidth]{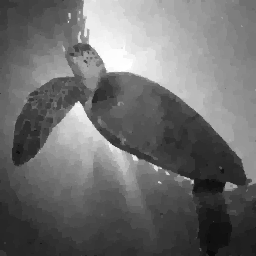}};
   		 \spy on (0.6, 2.3) in node [left] at (1.0, 0.5);
  	  \end{tikzpicture}
        \caption{Scalar TV 
        \\  $[29.17,  0.8035]$
        \newline}
    \end{subfigure}
    \hfill
    \begin{subfigure}[t]{0.24\textwidth}
             \begin{tikzpicture}[spy using outlines={rectangle, white, magnification=2, size=0.9cm, connect spies}]
  	  \node[anchor=south west,inner sep=0]  at (0,0) {
        \includegraphics[width=\textwidth]{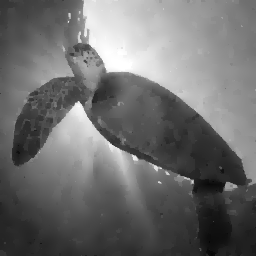}};
   		 \spy on (0.6, 2.3) in node [left] at (1.0, 0.5);
  	  \end{tikzpicture}
        \caption{Scalar TGV 
        \\ $[29.29,  0.8257]$
        \newline}
    \end{subfigure}
    \hfill \vspace{-0.2cm}
    \begin{subfigure}[t]{0.24\textwidth}
          \begin{tikzpicture}[spy using outlines={rectangle, white, magnification=2, size=0.9cm, connect spies}]
  	  \node[anchor=south west,inner sep=0]  at (0,0) {
        \includegraphics[width=\textwidth]{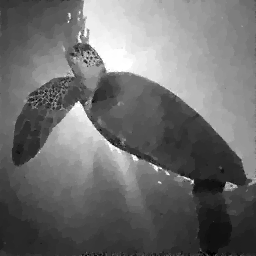}};
   		 \spy on (0.6, 2.3) in node [left] at (1.0, 0.5);
  	  \end{tikzpicture}
        \caption{WTV \cite{Calatroni_2019}\\ $[29.55,0.8137]$ \newline}
    \end{subfigure}
\hfill
    \begin{subfigure}[t]{0.24\textwidth}
             \begin{tikzpicture}[spy using outlines={rectangle, white, magnification=2, size=0.9cm, connect spies}]
  	  \node[anchor=south west,inner sep=0]  at (0,0) {
        \includegraphics[width=\textwidth]{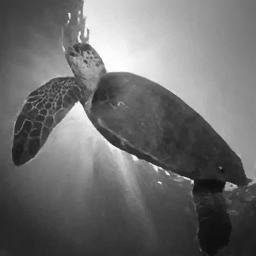}};
   		 \spy on (0.6, 2.3) in node [left] at (1.0, 0.5);
  	  \end{tikzpicture}
        \caption{U-TV \\ $[30.59,  0.8496]$ \newline}
    \end{subfigure}
    \hfill
  \begin{subfigure}[t]{0.24\textwidth}
         \begin{tikzpicture}[spy using outlines={rectangle, white, magnification=2, size=0.9cm, connect spies}]
  	  \node[anchor=south west,inner sep=0]  at (0,0) {
        \includegraphics[width=\textwidth]{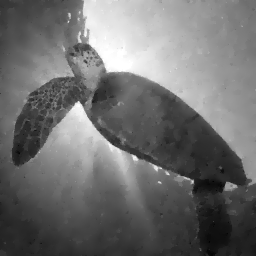}};
   		 \spy on (0.6, 2.3) in node [left] at (1.0, 0.5);
  	  \end{tikzpicture}
        \caption{Bilevel TGV \cite{bilevelTGV} \\ $[29.81, 0.8328]$ \newline}
    \end{subfigure}
    \hfill
    \begin{subfigure}[t]{0.24\textwidth}
       \begin{tikzpicture}[spy using outlines={rectangle, white, magnification=2, size=0.9cm, connect spies}]
  	  \node[anchor=south west,inner sep=0]  at (0,0) {
        \includegraphics[width=\textwidth]{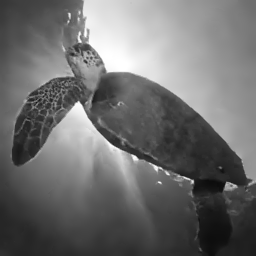}};
   		 \spy on (0.6, 2.3) in node [left] at (1.0, 0.5);
  	  \end{tikzpicture}
        \caption{U-TGV \\ $[\textbf{30.76},  \textbf{0.8600}]$ \newline}
    \end{subfigure}
    \hfill \vspace{-0.2cm}
    \begin{subfigure}[t]{0.24\textwidth}
        \includegraphics[width=\textwidth]{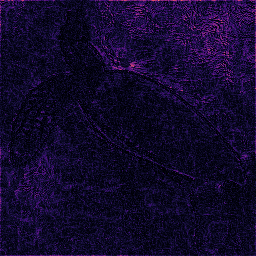}
        \caption{$\mathrm{\Lambda}$ map, U-TV}
    \end{subfigure}
    \hfill
    \begin{subfigure}[t]{0.24\textwidth}
        \includegraphics[width=\textwidth]{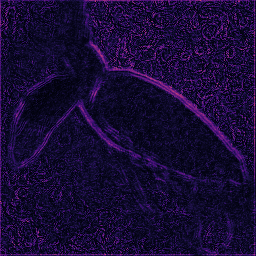}
        \caption{$\mathrm{\Lambda_1}$ map, U-TGV}
    \end{subfigure}
    \hfill
    \begin{subfigure}[t]{0.24\textwidth}
        \includegraphics[width=\textwidth]{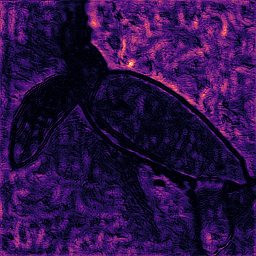}
        \caption{$\mathrm{\Lambda_0}$ map, U-TGV}
    \end{subfigure}
    \hfill
    \begin{subfigure}[t]{0.24\textwidth}
        \includegraphics[width=\textwidth]{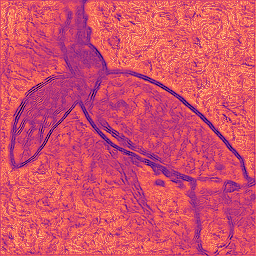}
    \caption{$\mathrm{\Lambda}_0/\mathrm{\Lambda}_1$ ratio}
    \end{subfigure}

    \begin{subfigure}[t]{1\textwidth}
        \centering
       \begin{tikzpicture} \node at (0,0) {
            \customcolorbar{magma_custom}{0}{0.85}{0.8\linewidth}{}{}
        }; \end{tikzpicture}
    \end{subfigure}

    \caption{Denoising results for the ``turtle'' image. The numbers is brackets show the [PSRN, SSIM] values. The colorbar is common among the visualisations of the parameter maps, except
    the $\mathrm{\Lambda}_0/\mathrm{\Lambda}_1$ ratio shown in logarithmic scale.   
    }
    \label{fig:turtle_test_case_noise_0_10}
\end{figure}
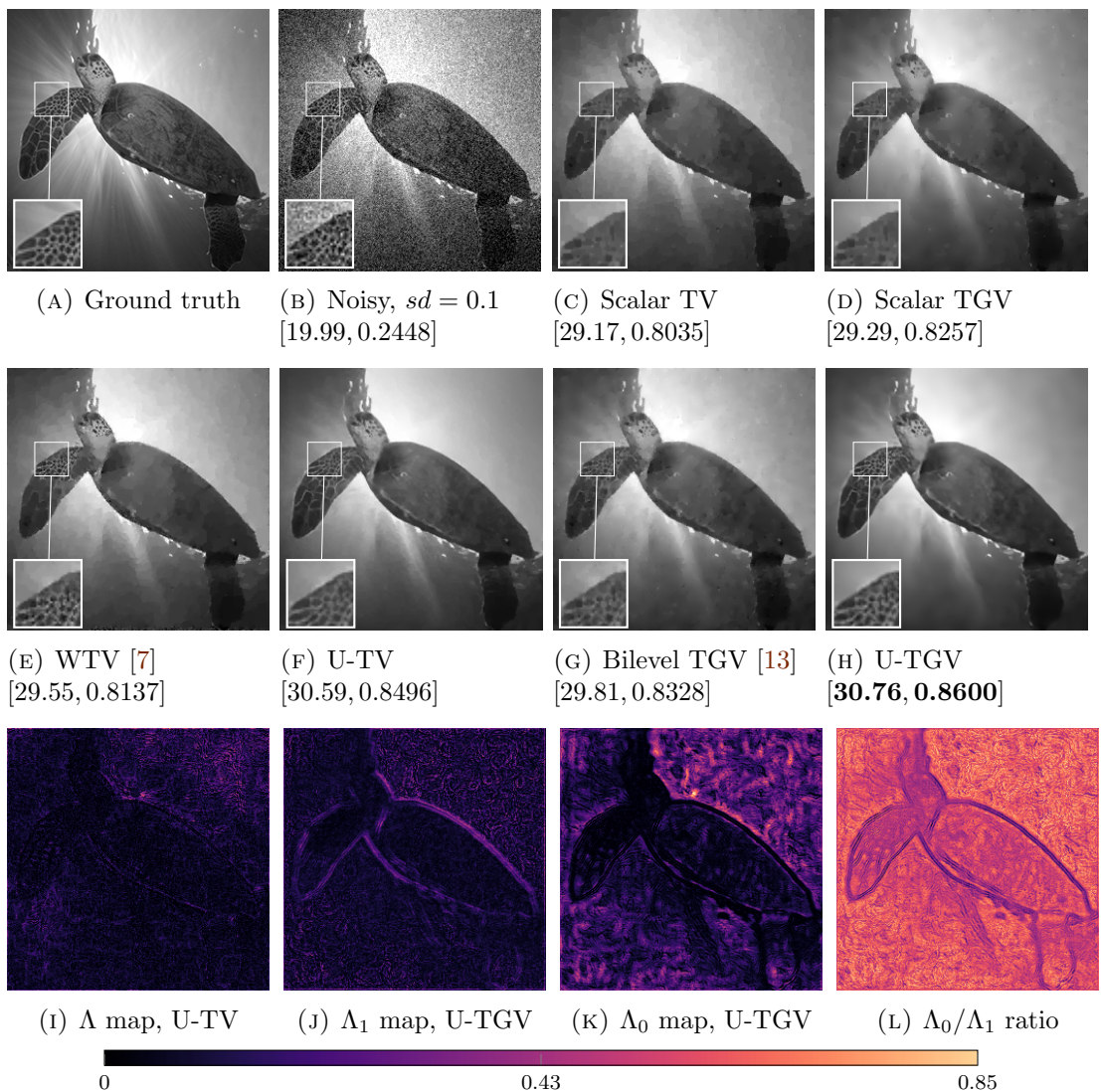

\begin{figure}[!t]
    \centering
    \begin{subfigure}[t]{0.24\textwidth}
        \begin{tikzpicture}[spy using outlines={rectangle, white, magnification=2, size=0.9cm, connect spies}]
  	  \node[anchor=south west,inner sep=0]  at (0,0) {
        \includegraphics[width=\textwidth]{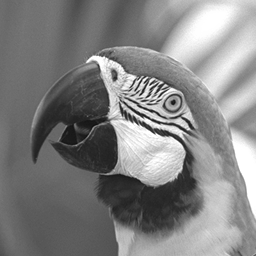}};
   		 \spy on (2.0, 2.1) in node [left] at (1.0, 0.5);
  	  \end{tikzpicture}
        \caption{Ground truth}
    \end{subfigure}
    \hfill
    \begin{subfigure}[t]{0.24\textwidth}
        \begin{tikzpicture}[spy using outlines={rectangle, white, magnification=2, size=0.9cm, connect spies}]
  	  \node[anchor=south west,inner sep=0]  at (0,0) {
        \includegraphics[width=\textwidth]{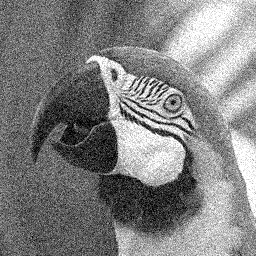}};
   		 \spy on (2.0, 2.1) in node [left] at (1.0, 0.5);
  	  \end{tikzpicture}
        \caption{Noisy, $sd=0.1$ \\  $[20.04, 0.2773]$ \newline}
    \end{subfigure}
    \hfill
    \begin{subfigure}[t]{0.24\textwidth}
        \begin{tikzpicture}[spy using outlines={rectangle, white, magnification=2, size=0.9cm, connect spies}]
  	  \node[anchor=south west,inner sep=0]  at (0,0) {
        \includegraphics[width=\textwidth]{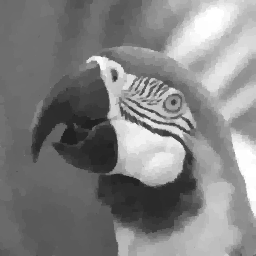}};
   		 \spy on (2.0, 2.1) in node [left] at (1.0, 0.5);
  	  \end{tikzpicture}
        \caption{Scalar TV 
        \\ $[28.59, 0.846]$
        \newline}
    \end{subfigure}
    \hfill
    \begin{subfigure}[t]{0.24\textwidth}
        \begin{tikzpicture}[spy using outlines={rectangle, white, magnification=2, size=0.9cm, connect spies}]
  	  \node[anchor=south west,inner sep=0]  at (0,0) {
        \includegraphics[width=\textwidth]{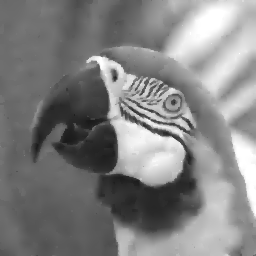}};
   		 \spy on (2.0, 2.1) in node [left] at (1.0, 0.5);
  	  \end{tikzpicture}
        \caption{Scalar TGV 
        \\ $[28.95, 0.8620]$
        \newline}
    \end{subfigure}
    \hfill \vspace{-0.2cm}
    \begin{subfigure}[t]{0.24\textwidth}
       	\begin{tikzpicture}[spy using outlines={rectangle, white, magnification=2, size=0.9cm, connect spies}]
  	  \node[anchor=south west,inner sep=0]  at (0,0) {
      	  \includegraphics[width=\textwidth]{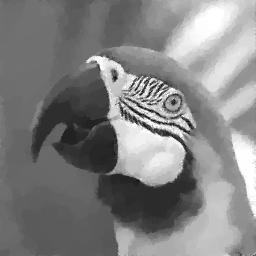}};
   		 \spy on (2.0, 2.1) in node [left] at (1.0, 0.5);
  	 \end{tikzpicture}
        \caption{WTV \cite{Calatroni_2019} \\ $[29.33, 0.8507]$}
    \end{subfigure}
    \hfill
    \begin{subfigure}[t]{0.24\textwidth}
           \begin{tikzpicture}[spy using outlines={rectangle, white, magnification=2, size=0.9cm, connect spies}]
  	  \node[anchor=south west,inner sep=0]  at (0,0) {
        \includegraphics[width=\textwidth]{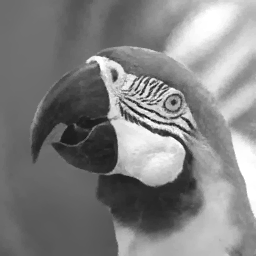}};
   		 \spy on (2.0, 2.1) in node [left] at (1.0, 0.5);
  	  \end{tikzpicture}
        \caption{U-TV \\ $[30.87, 0.8800]$ \newline}
    \end{subfigure}
    \hfill
    \begin{subfigure}[t]{0.24\textwidth}
       	  \begin{tikzpicture}[spy using outlines={rectangle, white, magnification=2, size=0.9cm, connect spies}]
  	  \node[anchor=south west,inner sep=0]  at (0,0) {
      	  \includegraphics[width=\textwidth]{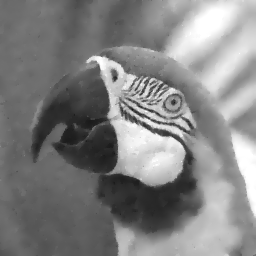}};
   		 \spy on (2.0, 2.1) in node [left] at (1.0, 0.5);
  	  \end{tikzpicture}
        \caption{Bilevel TGV \cite{bilevelTGV} \\  $[29.56, 0.8629]$ \newline}
    \end{subfigure}
    \hfill
    \begin{subfigure}[t]{0.24\textwidth}
          \begin{tikzpicture}[spy using outlines={rectangle, white, magnification=2, size=0.9cm, connect spies}]
  	  \node[anchor=south west,inner sep=0]  at (0,0) {
        \includegraphics[width=\textwidth]{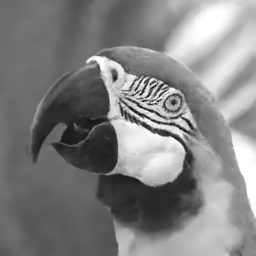}};
   		 \spy on (2.0, 2.1) in node [left] at (1.0, 0.5);
  	  \end{tikzpicture}
        \caption{U-TGV  \\ $[\textbf{31.03},\textbf{0.8883}]$ \newline}
    \end{subfigure}
    \hfill \vspace{-0.2cm}
    \begin{subfigure}[t]{0.24\textwidth}
        \includegraphics[width=\textwidth]{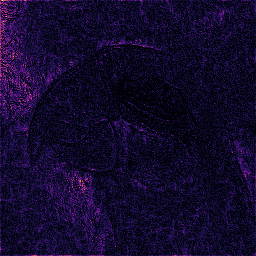}
        \caption{$\mathrm{\Lambda}$ map, U-TV \newline}
    \end{subfigure}
        \hfill
    \begin{subfigure}[t]{0.24\textwidth}
        \includegraphics[width=\textwidth]{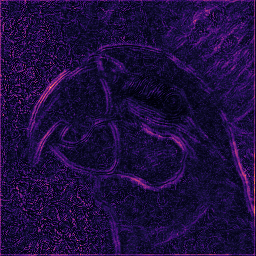}
        \caption{$\mathrm{\Lambda}_1$ map, U-TGV }
    \end{subfigure}
    \hfill
    \begin{subfigure}[t]{0.24\textwidth}
        \includegraphics[width=\textwidth]{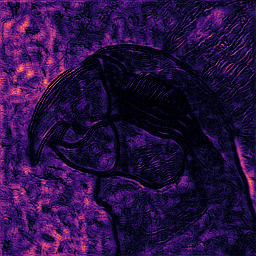}
        \caption{$\mathrm{\Lambda}_0$ map, U-TGV}
    \end{subfigure}
    \hfill
    \begin{subfigure}[t]{0.24\textwidth}
        \includegraphics[width=\textwidth]{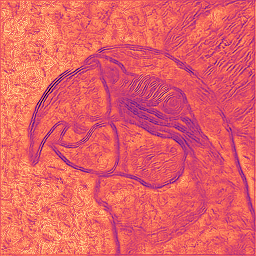}
        \caption{$\LL_0/\LL_1$ ratio 
        }
    \end{subfigure}
     \hfill \vspace{-0.3cm}
    \begin{subfigure}[t]{1\textwidth}
    \centering
         \begin{tikzpicture} \node at (0,0) {
            \customcolorbar{magma_custom}{0}{0.72}{0.8\linewidth}{}{}
        }; \end{tikzpicture}
    \end{subfigure}

    \caption{
    Denoising results for the ``parrot'' image. The numbers in brackets show the [PSRN, SSIM] values. The colorbar is common among the visualisations of the parameter maps, except
    the $\mathrm{\Lambda}_0/\mathrm{\Lambda}_1$ ratio shown in logarithmic scale. 
    }
    \label{fig:parrot_test_case_noise_0_10}
\end{figure}

Figures \ref{fig:turtle_test_case_noise_0_10} and \ref{fig:parrot_test_case_noise_0_10} show extensive results for the ``turtle'' and ``parrot'' images.
The ``turtle'' image belongs to the same distribution of images (the SeaTurtleID2022 dataset) the models were trained with, but was not part of the training set.
Apart from  U-TV and U-TGV, we also show the standard  TV and TGV results (best scalar parameter with respect to SSIM, found by a grid-search), as well the spatially varying bilevel TGV result of \cite{bilevelTGV} (unsupervised). We also report the spatially varying TV results (denoted by WTV) obtained by the algorithm proposed in \cite{Calatroni_2019} using a maximum-likelihood-type procedure (unsupervised).
We see that for both images U-TGV gives the best reconstruction, slightly better than U-TV, with both approaches significantly outperforming the other methods, both visually and quantitatively.
We also performed experiments with various noise levels in 50  natural images from
\cite{CCIA_dataset_UGR_dataset_2003}.
The results, shown in Table \ref{tab:summary_standard50_tests}, are aligned with the ones of the previous figures indicating that the models perform well in a wider distribution of images than the one trained on.

\subsection{Structure of the parameter-maps}\label{sec:triple_edge}

\begin{figure}[!t]
    \centering
    \begin{subfigure}[t]{0.24\textwidth}
        \includegraphics[width=\textwidth]{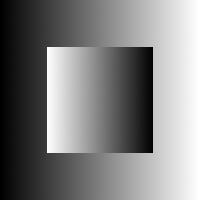}
 \caption{Ground truth}
    \end{subfigure}
    \hfill
    \begin{subfigure}[t]{0.24\textwidth}
        \includegraphics[width=\textwidth]{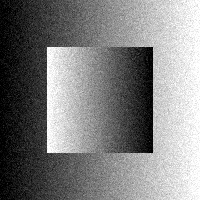}
 \caption{noisy, $sd=0.05$}
    \end{subfigure}
    \hfill
    \begin{subfigure}[t]{0.24\textwidth}
        \includegraphics[width=\textwidth]{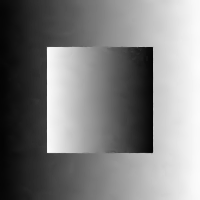}
\caption{U-TGV}
    \end{subfigure}
    \hfill
    \begin{subfigure}[t]{0.24\textwidth}
        \includegraphics[width=\textwidth]{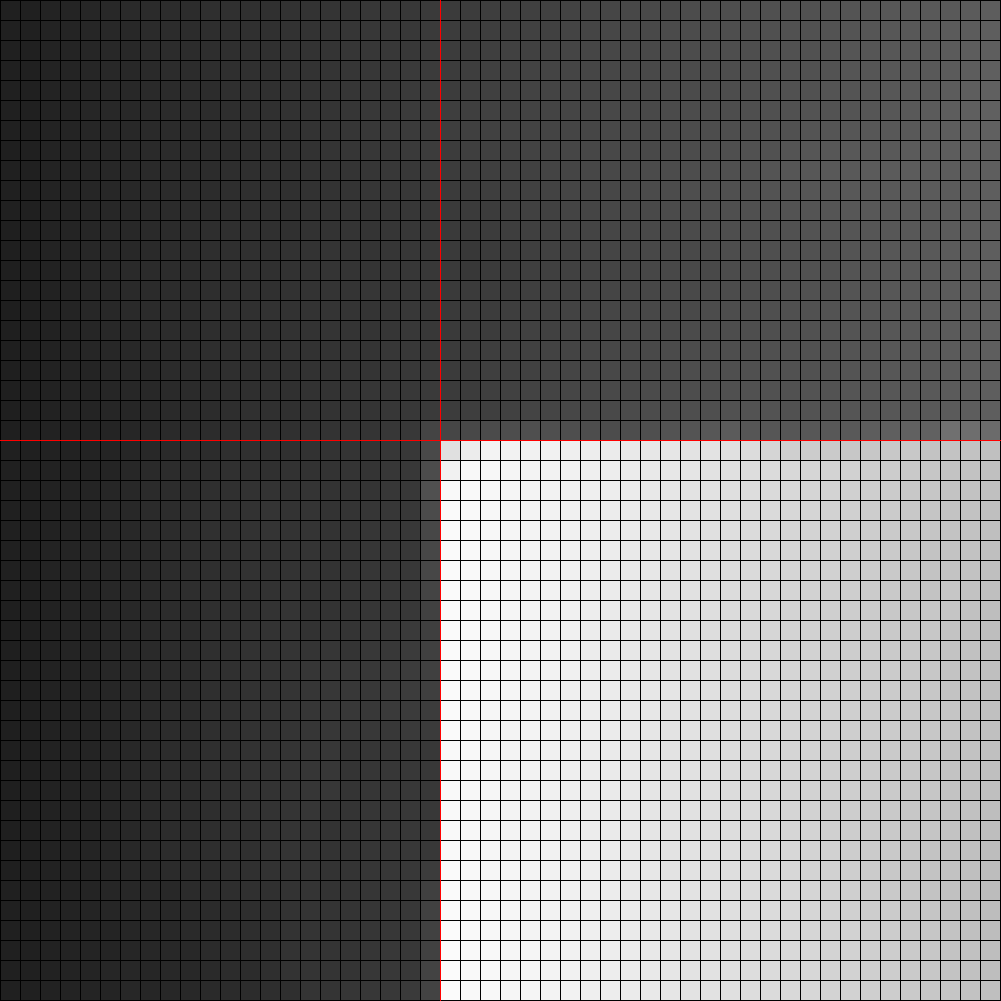}
        \caption{U-TGV, close-up}
    \end{subfigure}
  
    \begin{subfigure}[t]{0.24\textwidth}
        \includegraphics[width=\textwidth]{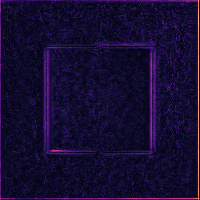}
       \caption{$\mathrm{\Lambda}_1$ map}
    \end{subfigure}
    \hfill
    \begin{subfigure}[t]{0.24\textwidth}
        \includegraphics[width=\textwidth]{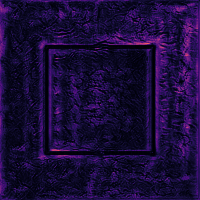}
        \caption{$\mathrm{\Lambda}_0$ map}
    \end{subfigure}
    \hfill
    \begin{subfigure}[t]{0.24\textwidth}
        \includegraphics[width=\textwidth]{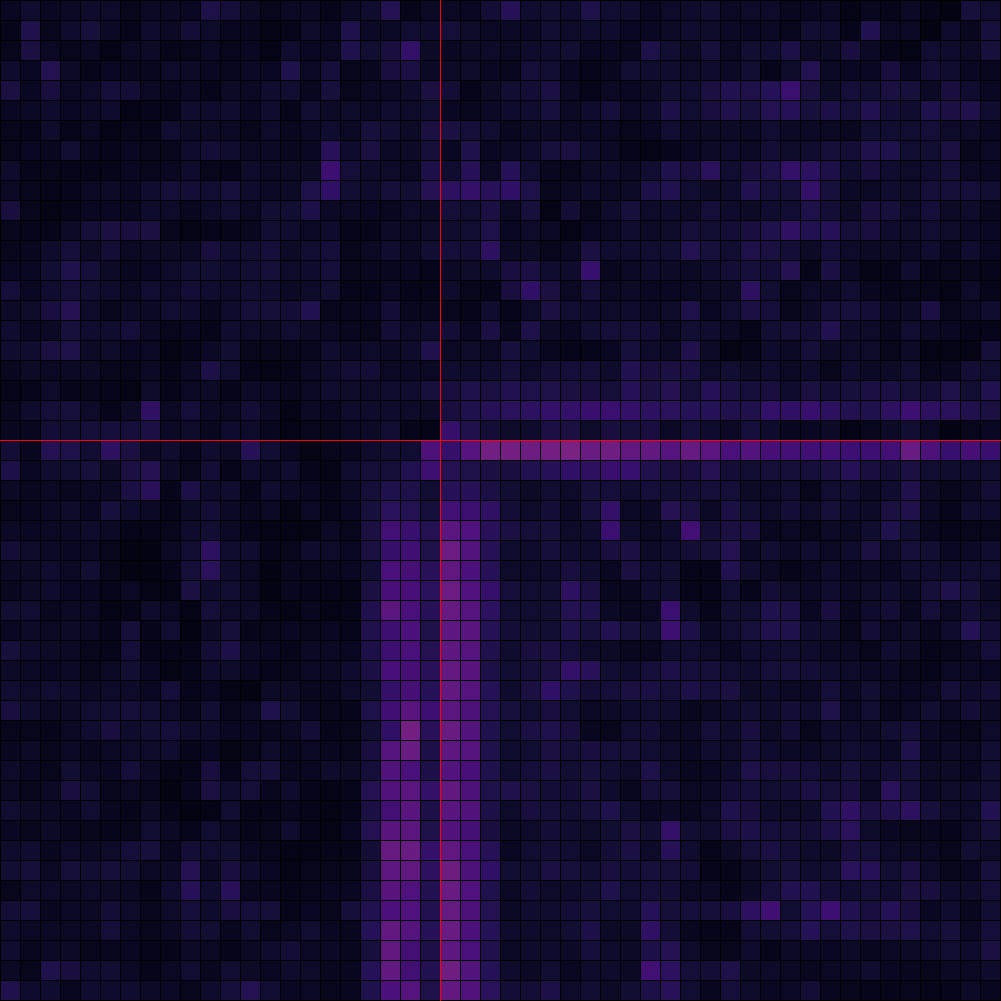}
        \caption{$\mathrm{\Lambda}_1$, close-up }
    \end{subfigure}
    \hfill
    \begin{subfigure}[t]{0.24\textwidth}
        \includegraphics[width=\textwidth]{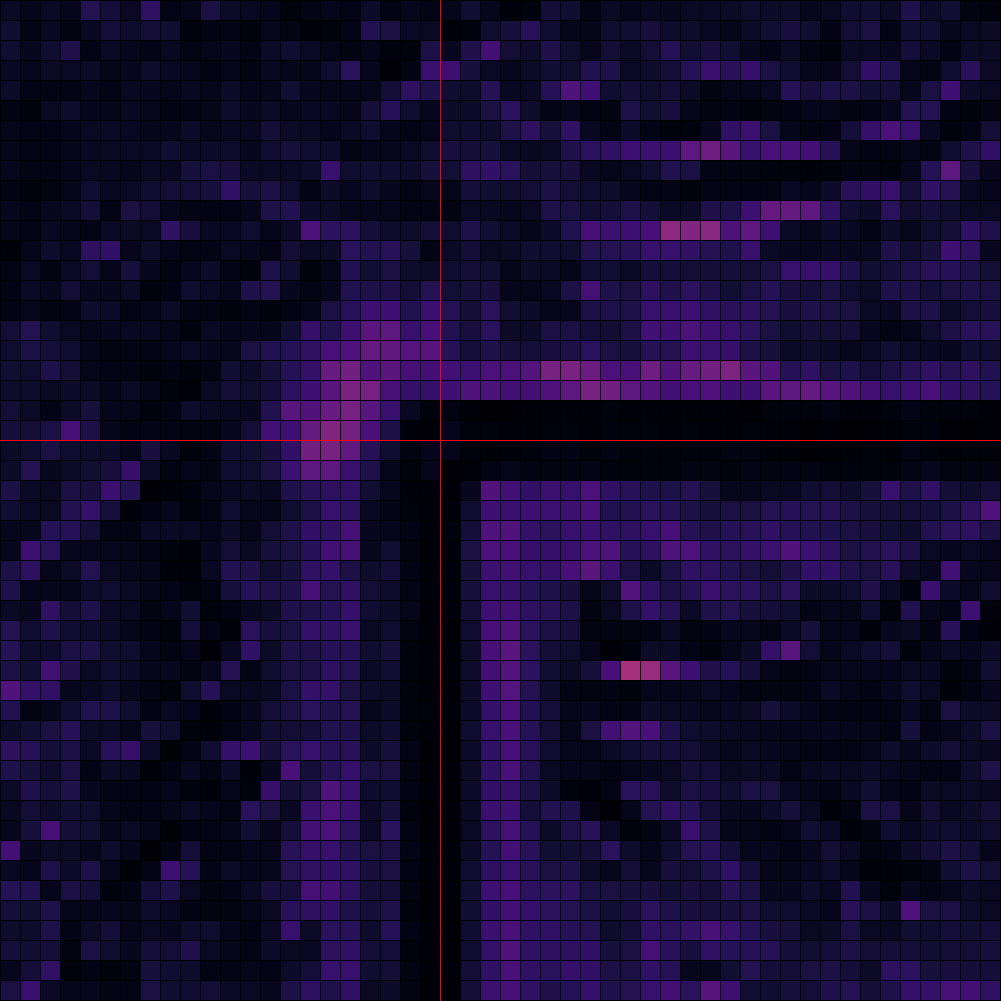}
        \caption{$\mathrm{\Lambda}_0$, close-up}
    \end{subfigure}

    \caption{Visualisation of the structure of the parameters $\mathrm{\Lambda}_0$ and $\mathrm{\Lambda}_1$ at the image edges. Both maps alternate between high-low-high values at the edges but $\mathrm{\Lambda}_0$ takes small values at a larger neighbourhood around the edges.
    }
    \label{fig:synthetic_square_triple_edge}
\end{figure}

Further insights are gained by examining the parameter maps in Figures \ref{fig:turtle_test_case_noise_0_10} and \ref{fig:parrot_test_case_noise_0_10}. By inspecting the ratio $\mathrm{\Lambda}_{0}/\mathrm{\Lambda}_{1}$ we can see the areas of the image where  TGV locally behaves like TV, c.f.\ Proposition \ref{prop:tgv_large_ratio}. This ratio is indeed higher in flatter, almost constant, areas.
A surprising observation is that the structure of $\LL$ for  U-TV is significantly different than the one of $\LL_{1}$ for U-TGV. While $\LL$ takes small values at edges and detailed areas of the image,  the values of $\LL_{1}$ alternate between being high-low-high at the edges. We also observe that the values of $\LL_{0}$ are small at a wider neighbourhood around the edges. To obtain a better visualisation of this \emph{triple-edge} phenomenon,  in Figure \ref{fig:synthetic_square_triple_edge}, we show the denoising results of the ``square'' image and the corresponding $\LL_{0}, \LL_{1}$ maps. We observe that $\LL_{1}$ takes small values exactly at the edges, resulting in small penalisation of the gradient there,  while $\LL_{0}$ takes small values at a wider area. Both parameter maps take larger values further away from the edges.

\begin{table}[!h]
    \centering
    \scriptsize 
    {\renewcommand{\arraystretch}{1.1}
    \begin{tabular}{c | c | c | c | c | c}
        $sd$ & metric & \textbf{scalar TV} & \textbf{scalar TGV} & \textbf{U-TV}  & \textbf{U-TGV} \\ \hline
        \textbf{0.05} & PSNR & 30.58 ± 1.78 & 30.81 ± 1.89 & 30.86 ± 2.70 & \textbf{31.03 ± 2.60} \\
        \textbf{} & SSIM & 0.8485 ± 0.033 & 0.8556 ± 0.033 & 0.8567 ± 0.051 &  \textbf{0.8590 ± 0.051} \\ \hline
        \textbf{0.1} & PSNR & 27.10 ± 2.19 & 27.36 ± 2.26 & 27.96 ± 2.67 & \textbf{28.11 ± 2.60} \\
        \textbf{} & SSIM & 0.7399 ± 0.058 & 0.7533 ± 0.056 & 0.7746 ± 0.068 &  \textbf{0.7786 ± 0.069} \\ \hline
        \textbf{0.15} & PSNR & 25.24 ± 2.37 & 25.49 ± 2.42 & 26.41 ± 2.67 &  \textbf{26.49 ± 2.61} \\
        \textbf{} & SSIM & 0.6668 ± 0.078 & 0.6822 ± 0.074 & 0.7135 ± 0.085 &  \textbf{0.7174 ± 0.086} \\ \hline
        \textbf{0.2} & PSNR & 23.89 ± 2.42 & 24.14 ± 2.46 & 25.38 ± 2.65 & \textbf{25.40 ± 2.58} \\
        \textbf{} & SSIM & 0.6133 ± 0.092 & 0.6286 ± 0.086 & 0.6686 ± 0.098 & \textbf{0.6706 ± 0.098} \\
    \end{tabular}
    }
    \vspace{0.5em}
    \caption{
        Denoising results for 50 natural images and for various noise levels.
    }
    \label{tab:summary_standard50_tests}
\end{table}

\section{Numerical experiments in MRI reconstruction}\label{sec:mri}
Here, we consider the case of accelerated MR image reconstruction, where the forward model in \eqref{intro:general_eq} is given by $A=P F$, where $F$ denotes the 2D Fourier transform and $P$ denotes a projection onto the set of the acquired measurements.\\[1em]
\subsection{Set-up}\label{sec:mri-set-up}
\begin{enumerate}[leftmargin=0pt]
 \item[] \emph{The training dataset:} The dataset consists of $M=3452$ pairs of retrospectively undersampled MR measurements and target images $(f^i,u_{\mathrm{true}}^{i})$ that are generated according to \eqref{intro:general_eq}, where the ground-truth images $u_{\mathrm{true}}^{i}$ are random samples extracted from 598 subjects from the fastMRI multi-coil brain dataset \cite{zbontar2018fastmri}. Since MR images are typically complex-valued and the fastMRI only provides root-sum-of-square (RSS) reconstructions as target images, we first estimated coil sensitivity maps from the fully sampled multi-coil measurements with \texttt{MRpro} \cite{zimmermann2025mrpro} using the method described in \cite{walsh2000adaptive}. Then, to obtain complex-valued images, we applied the adjoint of the multi-coil MR forward operator to the multi-coil measurement data and finally cropped the resulting images to a size of $320\times320$. We then retrospectively generated k-space data with random acceleration factor $R$ from $4, 5,\ldots, 8$ and adding zero-mean Gaussian noise with a random standard deviation in $[0, 0.2]$.  We used 3000, 150 and 302 images for training, validation and testing, respectively.

 \item[] \emph{The architecture of $\mathrm{NET}_{\theta}$:}
 The employed U-Net has a similar structure as before. 
 \item[] \emph{The choice of solution algorithm:}
As in the denoising case, 
we unrolled the PDHG algorithm as described in \cite{Knoll_2011, Kofler_2023}. This time we let the step sizes $\sigma, \tau$ to be trainable starting from $\sigma = \tau = 
1 / \sqrt{3}$, after which they took the values $\sigma=0.3414$, $\tau=0.3255$ for U-TV and $\sigma=0.1695$, $\tau=0.6553$ for U-TGV.
We set the extrapolation parameter $\theta=1$ for both models.
\item[] \emph{The choice of $N$:}
As before, we set $N=256$ for both models.
\item[] \emph{The training procedure:} We used the AdamW optimiser with a learning rate of $10^{-4}$, a weight decay  of $10^{-5}$ \cite{loshchilov2019decoupled}, a batch size of 1 and the MSE loss.
    Both models were trained for 100 epochs, and the epoch with the lowest validation error was chosen as the final model, $43^{\text{rd}}$  and $72^{\text{nd}}$ for U-TV and U-TGV respectively. Training took approximately 55 and 87 hours for U-TV and U-TGV respectively, done on the same machine as in denoising.
\end{enumerate}

\subsection{Results}

\begin{figure}[!t]
    \centering
    \begin{subfigure}[t]{0.24\textwidth}
      \begin{tikzpicture}[spy using outlines={rectangle, white, magnification=2, size=0.9cm, connect spies}]
  	  \node[anchor=south west,inner sep=0]  at (0,0) {
        \includegraphics[width=\textwidth]{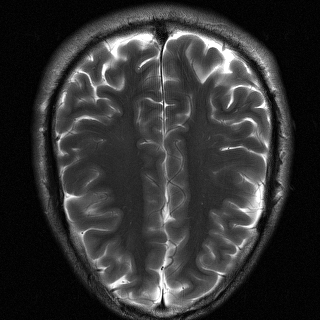}};
   		 \spy on (1.8, 0.4) in node [left] at (1.0, 0.5);
  	  \end{tikzpicture}
        \scriptsize
        \caption{Ground truth}
    \end{subfigure}
    \hfill
     \begin{subfigure}[t]{0.24\textwidth}
        \includegraphics[width=\textwidth]{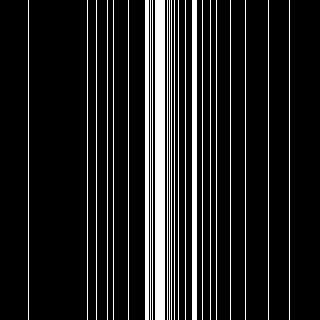}
        \scriptsize
        \caption{Sampling mask\\ with $R=8$
        \newline}
    \end{subfigure}
    \hfill
        \begin{subfigure}[t]{0.24\textwidth}
        \includegraphics[width=\textwidth]{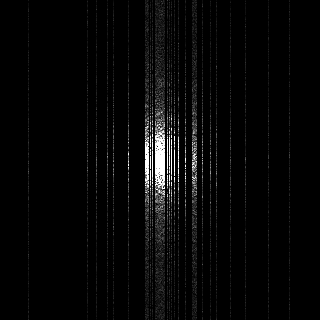}
        \scriptsize
        \caption{Noisy k-space \\ data, $sd=0.05$
        \newline}
    \end{subfigure}
    \hfill
    \begin{subfigure}[t]{0.24\textwidth}
     \begin{tikzpicture}[spy using outlines={rectangle, white, magnification=2, size=0.9cm, connect spies}]
  	  \node[anchor=south west,inner sep=0]  at (0,0){
        \includegraphics[width=\textwidth]{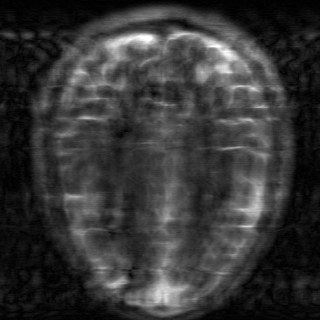}};
   		 \spy on (1.8, 0.4) in node [left] at (1.0, 0.5);
  	  \end{tikzpicture}
        \scriptsize
        \caption{Zero-filled\\  $[24.00, 0.6221]$
        \newline}
    \end{subfigure}
\hfill \vspace{-0.2cm}
        \begin{subfigure}[t]{0.24\textwidth}
        \begin{tikzpicture}[spy using outlines={rectangle, white, magnification=2, size=0.9cm, connect spies}]
  	  \node[anchor=south west,inner sep=0]  at (0,0){
        \includegraphics[width=\textwidth]{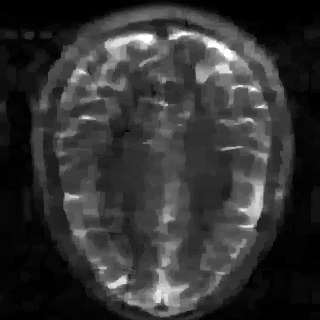}};
   		 \spy on (1.8, 0.4) in node [left] at (1.0, 0.5);
  	  \end{tikzpicture}
        \caption{Scalar TV
         \newline $[25.42, 0.6961]$
        \newline}
    \end{subfigure}
    \hfill
    \begin{subfigure}[t]{0.24\textwidth}
        \begin{tikzpicture}[spy using outlines={rectangle, white, magnification=2, size=0.9cm, connect spies}]
  	  \node[anchor=south west,inner sep=0]  at (0,0){
        \includegraphics[width=\textwidth]{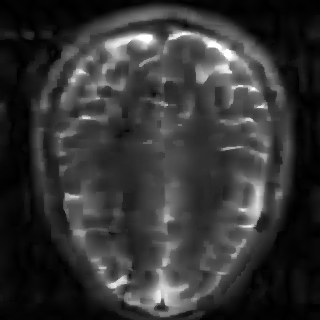}};
   		 \spy on (1.8, 0.4) in node [left] at (1.0, 0.5);
  	  \end{tikzpicture}
        \caption{Scalar TGV
        \newline $[25.65, 0.7204]$
        \newline}
    \end{subfigure}
    \hfill
    \begin{subfigure}[t]{0.24\textwidth}
         \begin{tikzpicture}[spy using outlines={rectangle, white, magnification=2, size=0.9cm, connect spies}]
  	  \node[anchor=south west,inner sep=0]  at (0,0){
        \includegraphics[width=\textwidth]{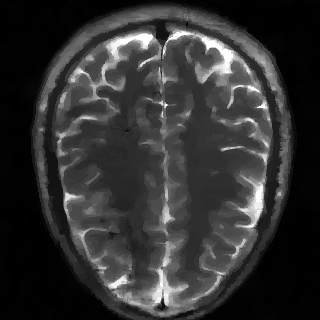}};
   		 \spy on (1.8, 0.4) in node [left] at (1.0, 0.5);
  	  \end{tikzpicture}
        \caption{U-TV\\ $[29.58, 0.7724]$
        \newline}
    \end{subfigure}
    \hfill
    \begin{subfigure}[t]{0.24\textwidth}
         \begin{tikzpicture}[spy using outlines={rectangle, white, magnification=2, size=0.9cm, connect spies}]
  	  \node[anchor=south west,inner sep=0]  at (0,0){
        \includegraphics[width=\textwidth]{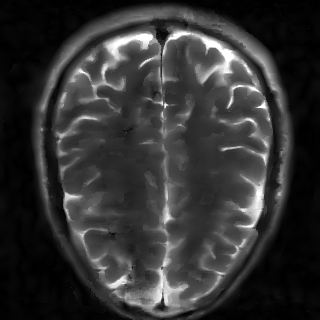}};
   		 \spy on (1.8, 0.4) in node [left] at (1.0, 0.5);
  	  \end{tikzpicture}
        \caption{U-TGV \\ $[\textbf{29.93},\textbf{0.8355}]$
        \newline}
    \end{subfigure}
    \hfill \vspace{-0.2cm}
      \begin{subfigure}[t]{0.24\textwidth}
        \includegraphics[width=\textwidth]{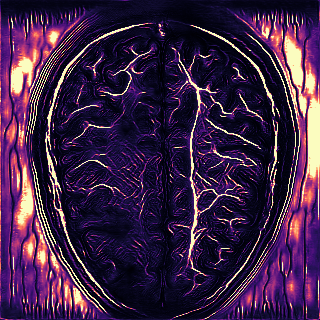}
        \begin{tikzpicture} \node at (0,0) {
            \customcolorbar{magma_custom}{0}{0.05}{0.75\linewidth}{$\leq$}{$\geq$}
        }; \end{tikzpicture}
         \captionsetup{font=small, skip=-3pt}
        \caption{$\LL$ map, U-TV}
    \end{subfigure}
    \hfill
     \begin{subfigure}[t]{0.24\textwidth}
        \includegraphics[width=\textwidth]{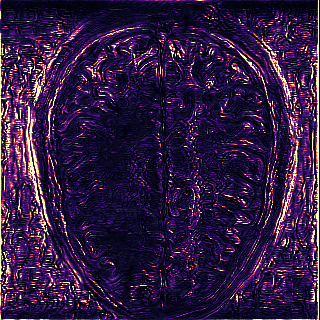}
        \begin{tikzpicture} \node at (0,0) {
            \customcolorbar{magma_custom}{0}{0.05}{0.75\linewidth}{$\leq$}{$\geq$}
        }; \end{tikzpicture}
         \captionsetup{font=small, skip=-3pt}
        \caption{$\LL_{1}$ map, U-TGV}
    \end{subfigure}
    \hfill
     \begin{subfigure}[t]{0.24\textwidth}
        \includegraphics[width=\textwidth]{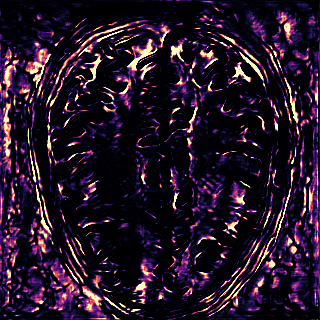}
        \begin{tikzpicture} \node at (0,0) {
            \customcolorbar{magma_custom}{0}{1}{0.75\linewidth}{$\leq$}{$\geq$}
        }; \end{tikzpicture}
         \captionsetup{font=small, skip=-3pt}
        \caption{$\LL_{0}$ map, U-TGV}
    \end{subfigure}
    \hfill
     \begin{subfigure}[t]{0.24\textwidth}
        \includegraphics[width=\textwidth]{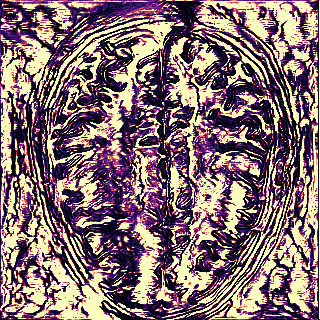}
        \begin{tikzpicture} \node at (0,0) {
            \customcolorbar{magma_custom}{0}{10}{0.75\linewidth}{$\leq$}{$\geq$}
        }; \end{tikzpicture}
         \captionsetup{font=small, skip=-3pt}
        \caption{$\LL_{0}/\LL_{1}$ ratio}
    \end{subfigure}
    \hfill

    \caption{MRI reconstruction results for a Brain image. The numbers in brackets show the [PSNR, SSIM] values. %
    }
    \label{fig:mri_results}
\end{figure}
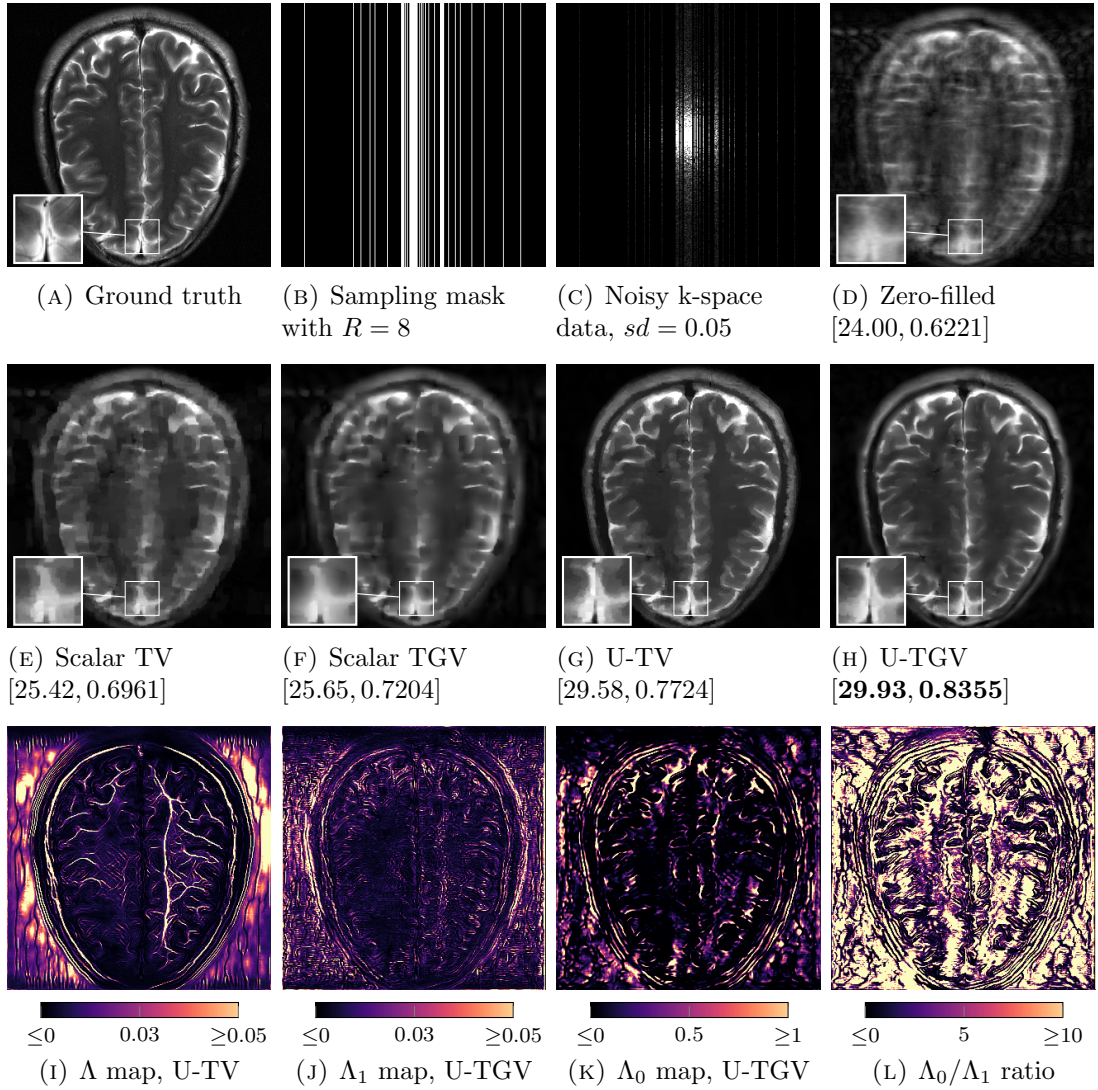

Figure \ref{fig:mri_results}
compares reconstructions for a test image under $R=8$, and Gaussian noise of $sd=0.05$. The best result is that of U-TGV,  with a significant margin over U-TV, particularly with respect to SSIM, and more than 4dB higher PSNR than the scalar version. We  observe again a  triple-edge structure for the parameter maps, with $\mathrm{\Lambda}_{0}$ taking small values in larger areas than $\mathrm{\Lambda}_{1}$. In Table \ref{tab:summary_MRI_tests}, we summarise the results for all  test images for $R=4,8$ and various noise levels. Apart from the low noise regime, U-TGV  performs slightly better than U-TV with respect to PSNR, but significantly better with respect to SSIM  in all cases. Both models heavily outperform the scalar versions.

 \begin{table}[!t]
  \centering
  \scriptsize
   {\renewcommand{\arraystretch}{1.1}
  \begin{tabular}{c|c|c|c|c|c|c|c}
      $R$ &$sd$ & metric   &  \textbf{zero-filled} & \textbf{Scalar TV} & \textbf{Scalar TGV} & \textbf{U-TV} & \textbf{U-TGV}\\ \hline

      \textbf{4} & \textbf{0.05} & PSNR & 24.93 ± 1.56 & 27.21 ± 1.81 & 28.58 ± 2.14 & \textbf{32.60} ± 2.30 & 32.43 ± 2.28 \\

      \textbf{ } & \textbf{}     & SSIM & 0.6361 ± 0.045 & 0.7134 ± 0.088 & 0.7619 ± 0.064 & 0.7129 ± 0.091 &  \textbf{0.8022} ± 0.079 \\ \hline

      \textbf{ } & \textbf{0.1}  & PSNR & 24.83 ± 1.55 & 27.15 ± 1.82 & 28.39 ± 2.12 & 32.01 ± 2.20 &  \textbf{32.05} ± 2.22 \\

      \textbf{ } & \textbf{}    & SSIM &  0.6235 ± 0.046  & 0.7144 ± 0.090 & 0.7590 ± 0.064 &  0.7039 ± 0.093 &  \textbf{0.7969} ± 0.079 \\ \hline

      \textbf{ } & \textbf{0.2}  & PSNR & 24.41 ± 1.53 & 26.91 ± 1.79 & 27.93 ± 2.09 & 30.94 ± 2.11 &  \textbf{31.23} ± 2.14  \\

      \textbf{ } & \textbf{} &SSIM  & 0.5607 ± 0.048 & 0.7096 ± 0.088 & 0.7515 ± 0.063 & 0.6976 ± 0.095 &  \textbf{0.7909} ± 0.077 \\ \hline

      \textbf{8} & \textbf{0.05} &PSNR & 23.08 ± 1.55 & 23.92 ± 1.71 & 25.02 ± 1.99 & \textbf{29.65} ± 2.25 & 29.62 ± 2.31 \\

      \textbf{ } & \textbf{} &SSIM & 0.5723 ± 0.053 & 0.6262 ± 0.085 & 0.6736 ± 0.070 &  0.6681 ± 0.095  &  \textbf{0.7611} ± 0.081 \\ \hline

      \textbf{ } & \textbf{0.1} & PSNR &23.02 ± 1.55 & 23.87 ± 1.71 & 24.93 ± 1.98 &  29.22 ± 2.19 & \textbf{29.36} ± 2.28 \\

      \textbf{ } & \textbf{} & SSIM &0.5648 ± 0.056& 0.6264 ± 0.085 & 0.6719 ± 0.069 & 0.6607 ± 0.095  & \textbf{0.7578} ± 0.080 \\ \hline

      \textbf{ } & \textbf{0.2} & PSNR &22.76 ± 1.54 & 23.77 ± 1.71 & 24.66 ± 1.94 & 28.38 ± 2.14 &  \textbf{28.73} ± 2.24 \\

      \textbf{ } & \textbf{} & SSIM &0.5065 ± 0.056  & 0.6249 ± 0.085 & 0.6676 ± 0.068 & 0.6608 ± 0.095 &  \textbf{0.7549} ± 0.077 \\
  \end{tabular}
  }
  \vspace{0.5em}
  \caption{Summary of MRI reconstruction results for all 302 test images.
  }
    \label{tab:summary_MRI_tests}
\end{table}

\section{Conclusion}\label{sec:conclusion}
We showed that using a CNN to infer spatially varying TGV regularisation parameters, which are highly adaptive to the data, significantly boosts its performance both in denoising and MRI reconstruction. The results enjoy high interpretability being solutions of variational problems with a handcrafted prior, as all the neural network-related ``black-boxness'' is transferred to the regularisation parameters rather than the reconstructed images themselves. Related to that, a further theoretical investigation of the structure of the parameter maps (triple-edge phenomenon) and their interplay in the regularisation process is of great interest.\\

\noindent
\textbf{Acknowledgments:}
We thank Luca Calatroni for producing the WTV results.

\bibliographystyle{amsplain}
\bibliography{refs}
\end{document}